%% file: kbc.tex
\documentclass{article}
\usepackage{aaai22}

\usepackage{times}  
\usepackage{helvet}  
\usepackage{courier}  
\usepackage[hyphens]{url}  
\usepackage{graphicx} 
\usepackage{natbib}  
\usepackage{caption} 
\DeclareCaptionStyle{ruled}{labelfont=normalfont,labelsep=colon,strut=off} 
\usepackage{booktabs}       
\usepackage{amsfonts,amsmath}       
\usepackage{nicefrac}       
\usepackage{microtype}      
\usepackage{xcolor}         
\usepackage{multirow}
\usepackage{graphicx}
\usepackage{tikz}

\input{math_commands.tex}

\newcommand{\eat}[1]{}
\newcommand{\mysecref}[1]{Section~\ref{#1}}
\newcommand{\mytabref}[1]{Table~\ref{#1}}
\newcommand{\myeqnref}[1]{Equation~\ref{#1}}

\newcommand{\myfigref}[1]{Figure~\ref{#1}}

\title{Combining Rules and Embeddings via Neuro-Symbolic AI for Knowledge Base Completion}
\author{Prithviraj Sen, Breno W. S. R. de Carvalho, Ibrahim Abdelaziz, Pavan Kapanipathi,\\Francois Luus, Salim Roukos, Alexander Gray}
\affiliations{IBM Research}

\newcommand{\squishlist}{
\begin{list}{$\bullet$}
{ \setlength{\itemsep}{0pt} \setlength{\parsep}{3pt}
\setlength{\topsep}{3pt} \setlength{\partopsep}{0pt}
\setlength{\leftmargin}{0em} \setlength{\labelwidth}{1em}
\setlength{\labelsep}{0.5em} } }

\newcommand{\squishlisttwo}{
\begin{list}{$\bullet$}
{ \setlength{\itemsep}{0pt} \setlength{\parsep}{0pt}
\setlength{\topsep}{0pt} \setlength{\partopsep}{0pt}
\setlength{\leftmargin}{1em} \setlength{\labelwidth}{1.5em}
\setlength{\labelsep}{0.5em} } }

\newcommand{\squishend}{
\end{list}  }

\begin{document}

\maketitle

\input{sections/abstract}
\input{sections/introduction_pavan}

\input{sections/preliminaries}

\input{sections/LNN_chain}
\input{sections/experiments}

\input{sections/related_work_2}
\input{sections/conclusion}

\bibliography{kbc}
\end{document}

%% file: math_commands.tex
\usepackage{amsmath,amsfonts,bm}

\def\eqref#1{equation~\ref{#1}}

\def\1{\bm{1}}

\def\rve{{\mathbf{e}}}

\def\rvr{{\mathbf{r}}}

\def\rvw{{\mathbf{w}}}
\def\rvx{{\mathbf{x}}}

\def\vzero{{\bm{0}}}
\def\vone{{\bm{1}}}

\DeclareMathAlphabet{\mathsfit}{\encodingdefault}{\sfdefault}{m}{sl}
\SetMathAlphabet{\mathsfit}{bold}{\encodingdefault}{\sfdefault}{bx}{n}

\def\gE{{\mathcal{E}}}

\def\gG{{\mathcal{G}}}

\def\gP{{\mathcal{P}}}

\def\gR{{\mathcal{R}}}

\def\gV{{\mathcal{V}}}



%% file: sections/abstract.tex
\begin{abstract}

\eat{
Recent interest in Knowledge Base Completion (KBC) has led to a plethora of approaches based on reinforcement learning, inductive logic programming and graph embeddings. With so many approaches, it can be difficult sometimes to compare these methods and identify their commonalities and differences. 
One of the main reasons that adds to this confusion comes from the fact that previous works have utilized distinct KBC evaluation metrics some of which can lead to overly optimistic evaluation. 
In this work, we start by re-evaluating a number of previously proposed KBC methods using a coherent set of metrics placing them on the same footing. We then propose simple approaches to learn rules for predicting missing edges in a knowledge graph. We employ logical neural networks, a particularly powerful, differentiable, framework for learning rules based on real-valued logic. Furthermore, by combining our learned rules with graph embeddings, we show that our approach achieves state-of-the-art performance on multiple standard KBC benchmarks. 
}

Recent interest in Knowledge Base Completion (KBC) has led to a plethora of approaches based on reinforcement learning, inductive logic programming and graph embeddings. In particular, rule-based KBC has led to interpretable rules while being comparable in performance with graph embeddings. Even within rule-based KBC, there exist different approaches that lead to rules of varying quality and previous work has not always been precise in highlighting these differences. Another issue that plagues most rule-based KBC is the non-uniformity of relation paths: some relation sequences occur in very few paths while others appear very frequently. In this paper, we show that not all rule-based KBC models are the same and propose two distinct approaches that learn in one case: 1) a mixture of relations and the other 2) a mixture of paths. When implemented on top of neuro-symbolic AI, which learns rules by extending Boolean logic to real-valued logic, the latter model leads to superior KBC accuracy outperforming state-of-the-art rule-based KBC by $2$-$10\%$ in terms of mean reciprocal rank. Furthermore, to address the non-uniformity of relation paths, we combine rule-based KBC with graph embeddings thus improving our results even further and achieving the best of both worlds.


\end{abstract}

%% file: sections/introduction_pavan.tex
\section{Introduction}
\label{sec:introduction}

A number of approaches have been proposed for knowledge base completion (KBC), a popular task that addresses the inherent incompleteness of Knowledge Graphs (KG)~\cite{bollacker2008freebase,rebele2016yago}. Compared to embeddings-based techniques~\cite{sun2019rotate,lacroix2018canonical}, rule learning techniques for KBC can handle cold-start challenges for new entities whose embeddings are not available (inductive setting) \citep{yang:nips17,sadeghian:neurips19,das:iclr18}, and learn multi-hop, human-interpretable, first order logic rules that enables complex reasoning over KGs. For example, the following rule (from Freebase) infers a person's nationality given her/his place of birth and its corresponding country:
\begin{equation*}
\forall P,N ~ \exists L\!\!: \text{nationality}(P, N) \!\leftarrow \text{bornIn}(P,L) \land~ \text{partOf}(L,N)
\end{equation*}

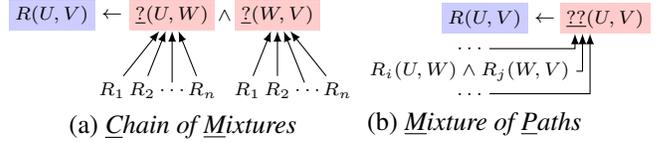
\begin{figure}
\begin{minipage}{0.55\linewidth}
{\scriptsize
\begin{tikzpicture}[>=latex]
 \node[rectangle,fill=blue!20] (rmix) at (0,0) {$R(U,V)$};
 \node (impl) at (0.8,0) {$\leftarrow$};
 \node[rectangle,fill=red!20] (fill1) at (1.6,0) {$\underline{?}(U,W)$};
 \node (conj) at (2.3,0) {$\wedge$};
 \node[rectangle,fill=red!20] (fill2) at (3,0) {$\underline{?}(W,V)$};
 
 \node (r1) at (0.8,-1) {$R_1$};
 \draw[->] (r1) -- (fill1);
 \node (r2) at (1.2,-1) {$R_2$};
 \draw[->] (r2) -- (fill1);
 \node (r3) at (1.6,-1) {$\ldots$};
 \draw[->] (r3) -- (fill1);
 \node (r4) at (2,-1) {$R_n$};
 \draw[->] (r4) -- (fill1);
 
 \node (r5) at (2.6,-1) {$R_1$};
 \draw[->] (r5) -- (fill2);
 \node (r6) at (3.0,-1) {$R_2$};
 \draw[->] (r6) -- (fill2);
 \node (r7) at (3.4,-1) {$\ldots$};
 \draw[->] (r7) -- (fill2);
 \node (r8) at (3.8,-1) {$R_n$};
 \draw[->] (r8) -- (fill2);
\end{tikzpicture}
}
\centerline{(a) \textit{\underline{C}hain of \underline{M}ixtures}}
\end{minipage}
\begin{minipage}{0.35\linewidth}
{\scriptsize
\begin{tikzpicture}[>=latex]
 \node[rectangle,fill=blue!20] (rmix) at (0,0) {$R(U,V)$};
 \node (impl) at (0.8,0) {$\leftarrow$};
 \node[rectangle,fill=red!20] (fill1) at (1.6,0) {$\underline{??}(U,V)$};
 
 \node (r2) at (-0.2,-0.4) {$\ldots$};
 \draw[->] (r2) -- (1.2,-0.4) -- (1.2,-0.2);
 \node (r3) at (-0.2,-0.7) {$R_i(U,W) \wedge R_j(W,V)$};
 \draw[->] (r3) -- (1.3,-0.7) -- (1.3,-0.2);
 \node (r4) at (-0.2,-1) {$\ldots$};
 \draw[->] (r4) -- (1.4,-1) -- (1.4,-0.2);
\end{tikzpicture}
}
\centerline{(b) \textit{\underline{M}ixture of \underline{P}aths}}
\end{minipage}
\caption{Rule-based KBC approaches: (a) CM, (b) MP.}
\label{fig:ruletemplates}
\end{figure}

\eat{The core ideas in Rule learning can be categorized into two groups based on their mechanism to select relations for rules. The first,  denoted as \textit{\underline{C}hain  of  Predicate  \underline{M}ixtures (CPM)}, represents each relation in the body of the rule as a mixture and has close ties to NeuralLP~\cite{neurallp}, DRUM~\cite{sadeghian2019drum}. The second, denoted as \textit{Mixture of Relation Paths (MRP)}, learns relation paths and has close ties to MINERVA~\cite{das:iclr18}, RNNLogic~\cite{qu2020rnnlogic}. Figure~\ref{fig:ruletemplates} depicts both CPM, MRP for learning a rule with two relations in the body. Recent trends in both these types of rule learning approaches has shown significant increase in complexity for performance gains over their simpler precursors. For instance: (a) DRUM, latest CPM approach, improves upon NeuralLP by proposing to learn multiple  rules each comprising mixtures of relations and also adds a low-rank tensor factorization scheme to keep the number of parameters in check. (b) RNNLogic, latest MRP approach, follows MINERVA in learning the relation sequences appearing in paths connecting source to destination vertices. However, it adds complexity by jointly training two modules\footnote{The joint training is done using expectation maximization}, a rule-generator for suggesting a set of high quality paths and a reasoning predictor that uses said paths to predict missing information.} 

There exist two kinds of prevalent rule learning approaches based on their mechanism to select relations. The first approach, denoted \textit{\underline{C}hain of \underline{M}ixtures} (CM),  represents each relation hop in the body of the rule, towards the right of the implication $\leftarrow$ symbol, as a mixture of relations. With close ties to NeuralLP~\citep{yang:nips17}, DRUM~\cite{sadeghian:neurips19}, CM is depicted in Figure~\ref{fig:ruletemplates} (a) where we learn a rule with two relations in the body. The second approach, denoted \textit{\underline{M}ixture of \underline{P}aths} (MP), learns relation paths with close ties to MINERVA~\cite{das:iclr18}, RNNLogic~\cite{qu2020rnnlogic}. Figure~\ref{fig:ruletemplates} (b) depicts MP where the body of the rule is defined as a mixture of all possible length $2$ relation paths in the KG. 
Crucially, both approaches, which rely on recurrent neural networks (RNN), use increasingly complex training algorithms. More precisely, NeuralLP and DRUM generate mixture probabilities using a long short-term memory (LSTM) \citep{hochreiter:neuralcomp97} and bi-directional LSTM, respectively. RNNLogic, the latest in the line of MP-based works, samples \emph{sets} of rules from a multinomial distribution defined by its RNN's latent vectors and learning this RNN requires sampling from intractable posteriors. While RNN training has improved significantly \citep{pascanu:icml13}, we are still faced with issues when these are used to learn long-range dependencies \citep{trinh:iclrw18}. Suffices to say that it is unclear whether the current RNN-based rule-based KBC models are in their simplest form.

Instead of following the current trend of devising increasingly complex rule-based KBC approaches, we ask whether it is possible to devise an approach without resorting to RNNs?
To this end, we propose to utilize Logical Neural Networks (LNN) \cite{riegel:arxiv20}, a member of Neuro-Symbolic AI family (NeSy), to devise both a CM-based (\textit{LNN-CM}) and a MP-based approach (\textit{LNN-MP}). 
LNN is an extension of Boolean logic to the real-valued domain and is differentiable thus enabling us to learn rules end-to-end via gradient-based optimization.
Another advantage of LNN is that while other members of NeSy harbor tenuous connections to Boolean logic’s semantics, e.g., \citet{dong:iclr19}, LNN shares strong ties thus making the learned rules fully interpretable. More importantly, our \emph{RNN-less} approach is conceptually simpler to understand and reason about than recent rule-based KBC approaches while still performing at par, if not better.


\eat{\begin{figure}
    \centering
    \includegraphics[width=0.99\linewidth]{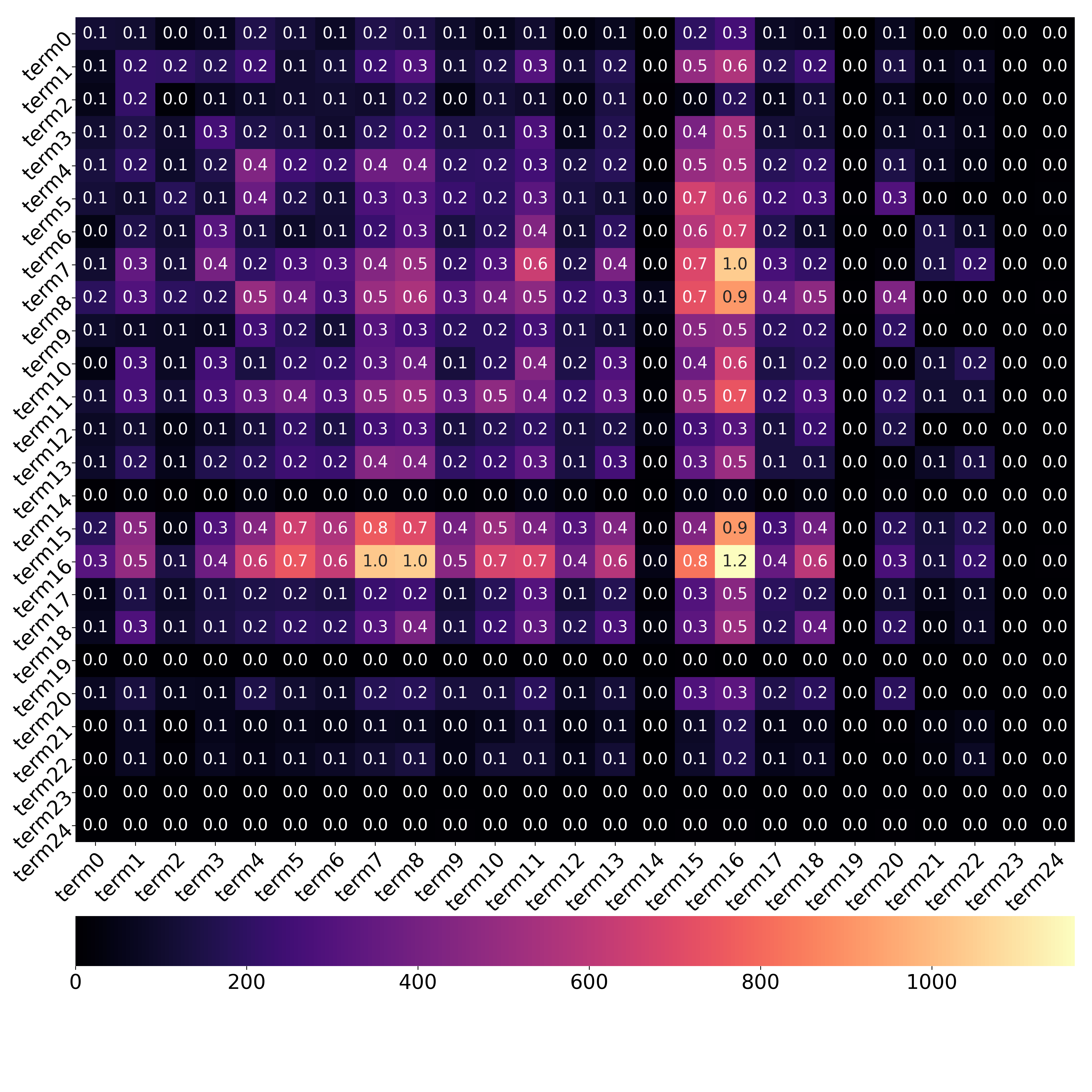}
    \caption{Path counts within cells are in thousands ($\times 1000$)}
    \label{fig:kinshipcounts}
\end{figure}}  
  
One shortcoming of MP (path-based rule learning) is that it suffers from sparsity. Since not all relation paths are equally prevalent in the KG, our goal is to guide the learning process towards more prevalent, effective relation paths as opposed to paths that appear less frequently. 
We show how to use pre-trained knowledge graph embeddings (KGE) to overcome such issues and learn more effective rules. While RNNLogic \citep{qu:iclr21} scores paths using RotatE~\cite{sun2019rotate}, a specific KGE, we show how to use any of the vast array of KGEs available in the literature. Our contributions are:

\eat{a rule consisting of a rare predicate $R$ (corresponding to fewer positive triples in the KB) simply will not have sufficient paths to reliably estimate model parameters. On the other hand, knowledge graph embeddings learn \emph{projection} operators \citep{ren:iclr20,ren:neurips20} based on neighborhood information which is abundant in comparison to paths between entities. However, combining rule-learning based approaches with such embedding-based techniques are rarely  been explored except~\citet{qu2020rnnlogic}. Therefore, inspired by~\citet{qu2020rnnlogic}{\color{red}Change this if necessary.}, we present a novel framework that can exploit the best of both rule-learning (LNN~\cite{riegel:arxiv20}) and embedding based approach (RotatE~\cite{sun2019rotate}). }

\squishlist
\item We propose to implement with logical neural networks (LNN) two simple approaches for rule-based KBC. LNN is a differentiable framework for real-valued logic that has strong connections to Boolean logic semantics.
\item To address the non-uniform distribution of relation paths that is likely to exist in any KG, we propose a simple approach that combines rule-based KBC with any KGE in order to reap the complementary benefits of both worlds.
\item Our experiments on $4$ different KBC benchmarks show that we: (a) outperform other rule-based baselines by 2\%-10\% (mean reciprocal rank); (b) are comparable to state-of-the-art rule with embedding-based technique RNNLogic.  
 \squishend

%% file: sections/preliminaries.tex
\section{Preliminaries: Logical Neural Networks}

We begin with a brief overview of \emph{Logical Neural Networks} (LNN) \citep{riegel:arxiv20}, a differentiable extension of Boolean logic that can learn rules end-to-end while still maintaining strong connections to Boolean logic semantics. In particular, LNN extends propositional Boolean operators with learnable parameters that allows a better fit to data. We next review operators such as LNN-$\wedge$, useful for modeling $\wedge$ in the example rule (\mysecref{sec:introduction}), and LNN-pred, useful for expressing mixtures used in both CM and MP (\myfigref{fig:ruletemplates}).

\subsection{Propositional LNN Operators}

To address the non-differentiability of classical Boolean logic, previous work has resorted to $t$-norms from fuzzy logic which are differentiable but lack parameters and thus cannot adapt to data. For instance, NeuralLP \citep{yang:nips17} uses product $t$-norm defined as $x \wedge y \equiv xy$ instead of Boolean conjunction. In contrast, LNN conjunction includes parameters that can not only express conjunctive semantics over real-valued logic but also fit the data better. The following LNN-$\wedge$ operator extends the \L ukasiewicz $t$-norm, defined as $x \wedge y \equiv \max(0, x+y-1)$, with parameters:

\begin{figure}
\centerline{
\begin{minipage}{0.49\linewidth}
\includegraphics[width=0.99\linewidth]{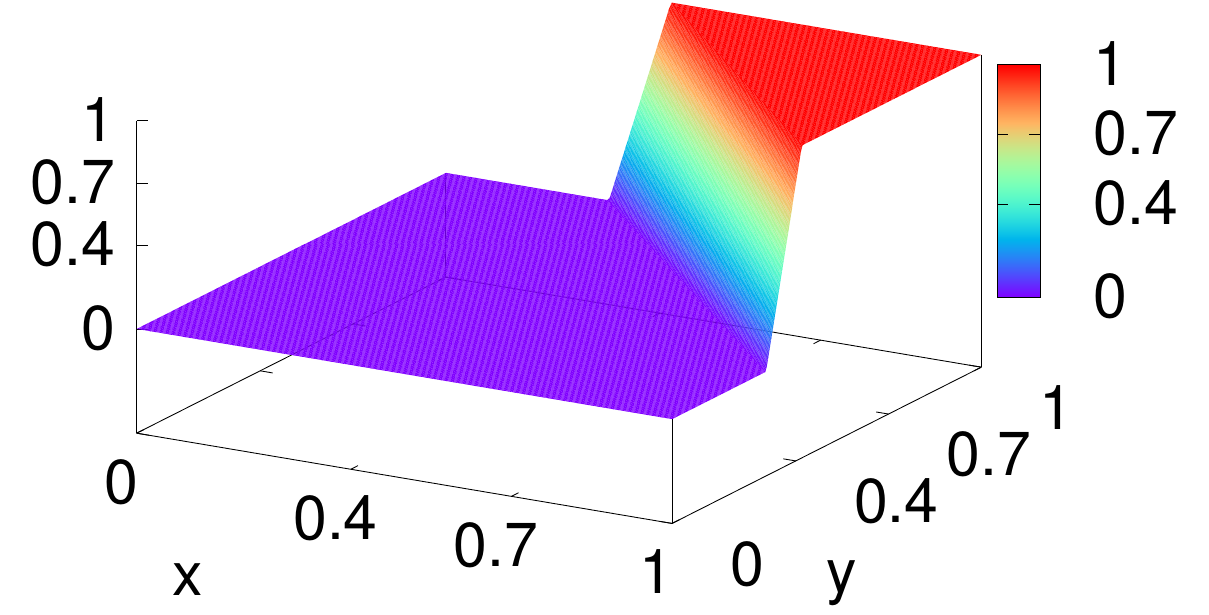}
\centerline{(a) LNN conjunction}
\end{minipage}
\hfill
\begin{minipage}{0.49\linewidth}
\includegraphics[width=0.99\linewidth]{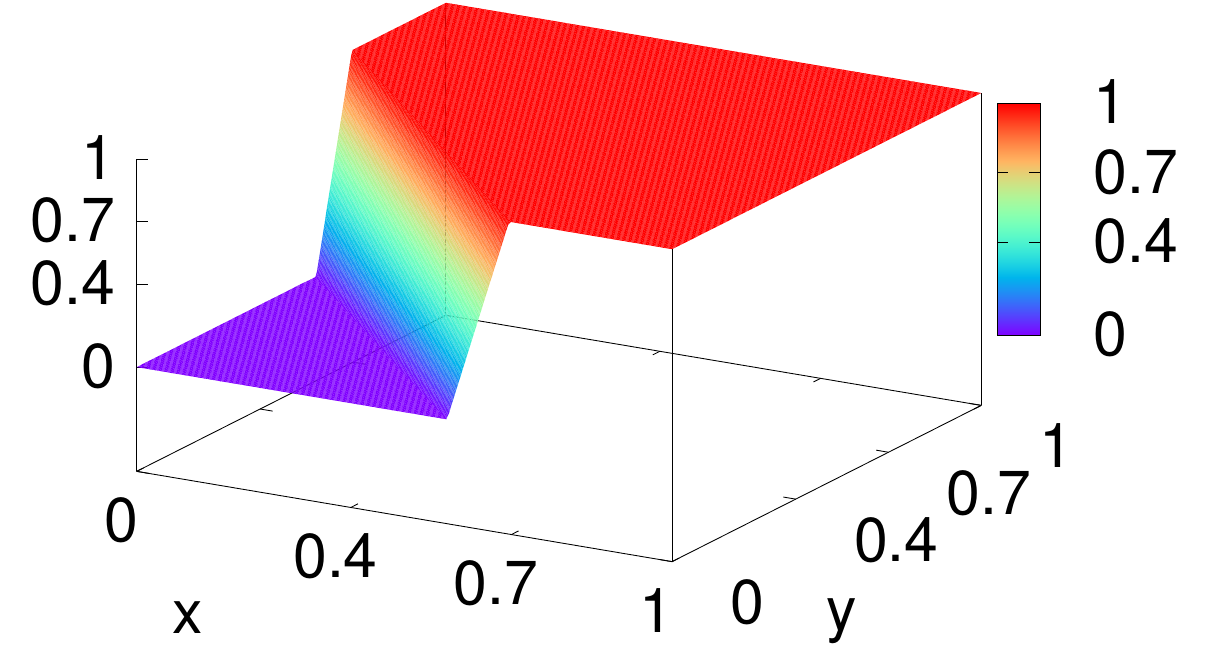}
\centerline{(b) LNN disjunction}
\end{minipage}}
\caption{(a) LNN-$\wedge$ and (b) LNN-$\vee$ learned with $\alpha=0.7$.}
\label{fig:lnn_ops}
\end{figure}

\begin{eqnarray}
\nonumber \text{LNN-}\!\wedge\!(\rvx;\beta,\rvw)\equiv&\max[0, \min\{1, \beta - \rvw^\top(\vone-\rvx)\}]&\\
\label{eqn:lowconstraint}\text{subject to:}&\beta\vone - \alpha \rvw \leq (1-\alpha)\vone&\\
\label{eqn:highconstraint}&\beta - (1-\alpha) \vone^\top \rvw \geq \alpha&\\
\nonumber&\rvw \geq \vzero&
\label{eqn:constraints}
\end{eqnarray}
where $\rvw$ denotes weights and $\beta$ denotes bias, both learnable parameters, $\rvx$ denotes a vector of inputs (in the case of \emph{binary} conjunction $|\rvx|=2$), $\vzero$ ($\vone$) denote a vector of 0s (1s), and $\alpha$ denotes a hyperparameter. Recall that, a (binary) Boolean conjunction operator returns {\tt true} ($1$) if both inputs are {\tt true} ($1$) and {\tt false} ($0$) otherwise. Intuitively, output is ``high" if both inputs are ``high", otherwise the output is ``low". There are $4$ steps to understanding how LNN extends this behavior to real-valued logic. 1) Given variable $x$ whose truth value lies in the range $[0,1]$, LNN utilizes $\alpha > \frac{1}{2}$ to define ``low" as $x \in [0, 1-\alpha]$ and ``high" as $x \in [\alpha, 1]$. 2) Note that, the non-negativity constraint enforced on $\rvw$ in LNN-$\wedge$ ensures that it is a monotonically increasing function with respect to input $\rvx$. 3) Given its monotonicity, we can ensure that LNN-$\wedge$ returns a ``high" value when all entries in $\rvx$ are ``high" by simply constraining the output to be ``high" when $\rvx = \alpha \vone$ (\myeqnref{eqn:highconstraint}). 4) Similarly, to ensure that the output is ``low'' if any input entry is ``low'', i.e. $1-\alpha$, \myeqnref{eqn:lowconstraint} constrains the output to be low when all but one entries in $\rvx$ is $1$. This constrained formulation of LNN-$\wedge$ ensures that conjunctive semantics is never lost during the learning process while still providing a better fit. 

\par

\myfigref{fig:lnn_ops} (a) shows a learned (binary) LNN-$\wedge$ ($\alpha=0.7$). Notice how, the output ($z$-axis) is close to $0$ when either input $x$ or $y$ is ``low" ($\in [0, 0.3]$) and jumps to $1$ when both are ``high" ($\in [0.7, 1]$), thus precisely capturing the semantics of logical conjunction. Note that, $\max$ and $\min$ have subgradients available thus making LNN-$\wedge$ amenable to training via gradient-based optimization. Just to contrast with LNN disjunction ($\vee$) that is defined as $1 - \text{LNN-}\wedge(1-\rvx;\beta,\rvw)$, \myfigref{fig:lnn_ops} (b) shows a learned LNN-$\vee$ ($\alpha=0.7$). In contrast to LNN-$\wedge$, in this case, output is ``low" when both $x, y \in [0,0.3]$ and jumps to $1$ when either of them are ``high" ($\in [0.7, 1]$) thus capturing semantics of Boolean disjunction.

Another operator we will find useful is LNN-$\text{pred}$. In the next section, we show how to use this to express mixtures, either over relations (CM) or relation paths (MP). LNN-pred is a simple operator with one non-negative weight per input:
\begin{equation*}
\text{LNN-}\text{pred}(\rvx; \rvw) = \rvw^\top\rvx ~ \text{subject to }\rvw \geq \vzero, \rvw^\top \vone = 1
\end{equation*}
where $\rvx$ denotes (a vector of) inputs and $\rvw$ denotes non-negative, learnable parameters constrained to sum to $1$.

\subsection{Training LNNs}

While LNN operators are amenable to gradient-based learning, one complication we are yet to address is implementing \emph{constrained} optimization to learn LNN parameters. Fortunately, there exist approaches that can convert any system of linear constraints (including equalities and inequalities) into a sequence of differentiable operations such that we can sample parameters directly from the feasible set \citep{frerix:cvprw20} which is what we use to train all our LNNs. We also refer the interested reader to \citet{riegel:arxiv20} that describes additional LNN training algorithms.

%% file: sections/LNN_chain.tex
\section{Learning LNN Chain Rules for KBC}
\label{sec:chainrules}

\subsection{Notation and Problem Definition}

Let $\gG = \langle \gV, \gR, \gE \rangle$ denote a knowledge graph (KG) such that edge $e \in \gE$ comprises a triple $\langle h, r, t \rangle$ where $h, t \in \gV$ denote source and destination vertices, and $r \in \gR$ denotes a relation. In this work, we focus on predicting destinations, i.e., given \emph{query} $\langle h, r, ?\rangle$ predict the \emph{answer} $t$. Following previous work \citep{yang:nips17}, we learn \emph{chain} rules in first-order logic (also called open path rules) that chains together multiple relations from $\gR$ to model the given relation: 
\begin{multline}
\forall X_0, X_m ~ \exists X_1, \ldots X_{m-1}:\\
r_0(X_0, X_m) \leftarrow r_1(X_0, X_1) \wedge \ldots \wedge r_m(X_{m-1}, X_m)\\
r_i \in \gR ~ \forall i=0, \ldots m
\end{multline}
where the \emph{head} $r_0$ denotes the relation being modeled, $r_1, \ldots r_m$ form the relations in the \emph{body}, and $X_0, \ldots X_m$ denote logical constants that can take values from $\gV$. Note that, relations can repeat within the body, i.e., $r_i = r_j, i \neq j$. Furthermore, we allow $r_0$ to appear within the body which leads to a recursive rule. Chain rules are closely connected to multi-hop paths. Essentially, the above rule claims that $\langle u, r_0, v \rangle$ exists \emph{if} there exists an $m$-length path $p=u \overset{r_1}{\longrightarrow} \ldots \overset{r_i}{\longrightarrow} \ldots \overset{r_m}{\longrightarrow} v$ connecting $X_0=u$ and $X_m=v$. Given such a multi-hop path $p$, we refer to the sequence of relations $r_1, \ldots r_m$ as its \emph{relation path}. Furthermore, given an $m$-length relation path $\rvr = r_1, \ldots r_m$, $\gP_{\rvr}(u, v)$ denotes the \emph{set} of all paths connecting $u$ to $v$ in $\gG$ via relation path $\rvr$.

\par

Our goal is to learn to predict destinations for each $r \in \gR$, however, in interest of keeping the approach simple we pose each relation-specific learning task in isolation. Given that standard KBC benchmarks (such as WN18RR) consist of sparse graphs, following previous work \citep{yang:nips17}, we too introduce inverse relations. In other words, for each $r \in \gR$ we introduce a new relation $r^{-1}$ by adding for each $\langle h, r, t \rangle \in \gE$ a new triple $\langle t, r^{-1}, h \rangle$ to $\gG$. We refer to the augmented set of relations, including the inverse relations, as $\gR^{+}$. Learning to predict destinations for a given relation $r \in \gR^{+}$ is essentially a binary classification task where $\langle h, r, t \rangle \in \gE$ and $\langle h, r, t \rangle \notin \gE, \forall h, t \in \gV$, denote positive and negative examples, respectively. The approaches we present next differ in how they score a triple $\langle h, r, t \rangle$ which in turn depends on paths $\gP_{\rvr}(u, v)$. Since we are interested in using such models to score unseen triples, we need to remove the triple from $\gG$ during training because, as mentioned earlier, we use triples from $\gG$ as positive examples. More precisely, when scoring a triple $\langle h, r, t \rangle$ during training we always compute $\gP_{\rvr}(u, v)$ not from $\gE$ but from $\gE \setminus \{\langle h, r, t \rangle, \langle t, r^{-1}, h \rangle\}$, i.e., we always remove $2$ edges: the triple itself and its corresponding inverse triple. 

\subsection{Chains of LNN-pred Mixtures}
\label{sec:CPM}
Given $\gG$, a relation to model $r \in \gR$, and a user-defined rule-length $m$, our first approach for learning chain rules uses $m$ LNN-pred operators and one $m$-ary LNN-$\wedge$ operator. Score for triple $\langle u, r, v \rangle$ is defined as:
\begin{equation*}
\mathop{\sum_{\rvr =}}_{r_1,\ldots,r_m} \mathop{\sum_{p \in}}_{\gP_{\rvr}(u,v)} \text{LNN-}\wedge\{\text{LNN-pred}(\rve_{r_1}), \ldots \text{LNN-pred}(\rve_{r_m})\}
\end{equation*}
where $\rve_r \in \{0, 1\}^{|\gR|}$ is a one-hot encoding whose $r^{th}$ entry is $1$ with $0$s everywhere else. To be clear, the above model consists of $m+1 + m|\gR|$ parameters: $m+1$ for LNN-$\wedge$ ($\rvw, \beta$) and $m$ LNN-pred operators each consisting of $|\gR|$-sized $\rvw$ parameters. Due to one-hot encodings and the fact that the inner summation term does not depend on path $p$, the above expression can be simplified to:
\begin{equation}
\sum_{\rvr =r_1, \ldots, r_m} |\gP_{\rvr}(u,v)| \text{LNN-}\wedge(w^1_{r_1}, \ldots w^m_{r_m})
\label{eqn:cm}
\end{equation}
where $w^i_r$ denotes the $r^{th}$ weight of the $i^{th}$ LNN-pred.

\par
 
This model is closely related to  NeuralLP \citep{yang:nips17} that also chains together mixtures of relations (see Equation 5 in \citeauthor{yang:nips17}). Differences between the above model and NeuralLP include the use of 1) LNN operators instead of product $t$-norm and, 2) a rule generator. NeuralLP uses an RNN-based rule generator to generate the relations present in the body of the chain rule, we instead sum over all relation paths. NeuralLP was among the first NeSy-based KBC approaches on top of which later approaches are built, e.g. DRUM \citep{sadeghian2019drum}.

\begin{figure}
    \centering
    \includegraphics[width=0.99\linewidth,trim={0cm 3.6cm 0cm 0cm},clip]{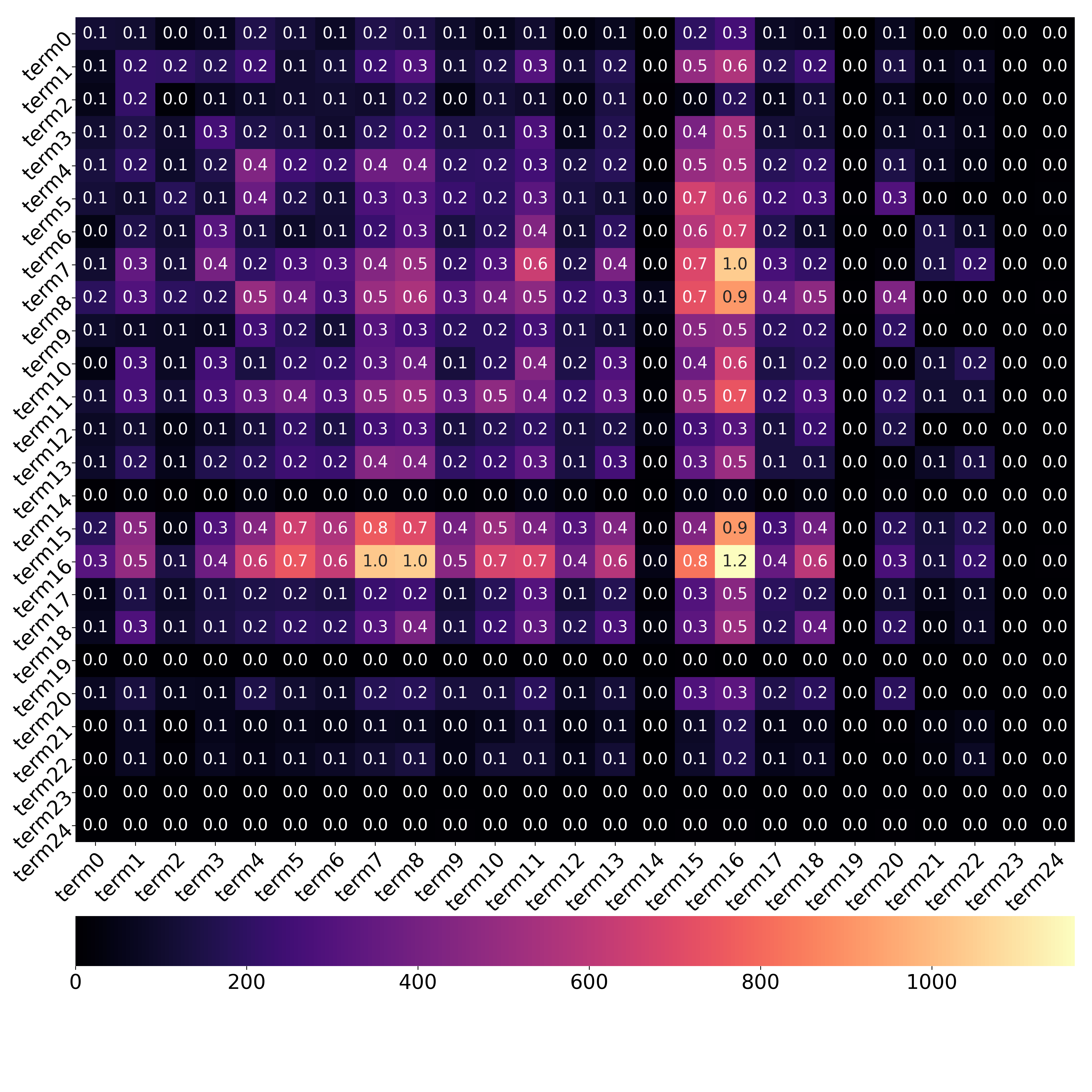}
    \caption{Kinship path counts (in thousands).}
    \label{fig:kinshipcounts}
\end{figure}

\subsection{Mixture of Relation Paths}
\label{sec:MRP}
One of the drawbacks of the previous model is that it treats each LNN-pred independently. To get around this, we propose our second approach which consists of one LNN-pred \emph{across all relation paths}. Given $\gG, r, m$, score for $\langle u, r, v \rangle$ is:
\begin{equation*}
\sum_{\rvr =r_1, \ldots, r_m}  |\gP_{\rvr}(u,v)| \text{LNN-pred}(\rve_{\rvr})
\end{equation*}
where $\rve_{\rvr} \in \{0, 1\}^{|\gR|^m}$ is a one-hot encoding with $1$ in its $\rvr^{\text{th}}$ position and $0$ everywhere else. One way to define a unique index for relation path $\rvr=r_1,\ldots r_m$ is $\sum_{i=1,\ldots m} r_i |\gR|^i$. 

\par

While recent approaches \citep{das:iclr18,qu:iclr21} have followed a similar model, the idea was proposed in the path ranking algorithm (PRA) \citep{lao:emnlp11}. Instead of path counts $|\gP_{\rvr}(u,v)|$, PRA computes the random walk probability of arriving from $u$ to $v$. Another difference lies in the parameterization. While LNN-pred constrains its parameters to sum to $1$, PRA uses elastic net regularization instead . Note that, PRA has been shown to perform quite poorly on KBC benchmarks, e.g. see \citet{qu:iclr21}.

\subsection{Handling Sparsity with Graph Embeddings}

\begin{figure}
    \centering
    \includegraphics[width=0.99\linewidth,trim={0cm 3.6cm 0cm 0cm},clip]{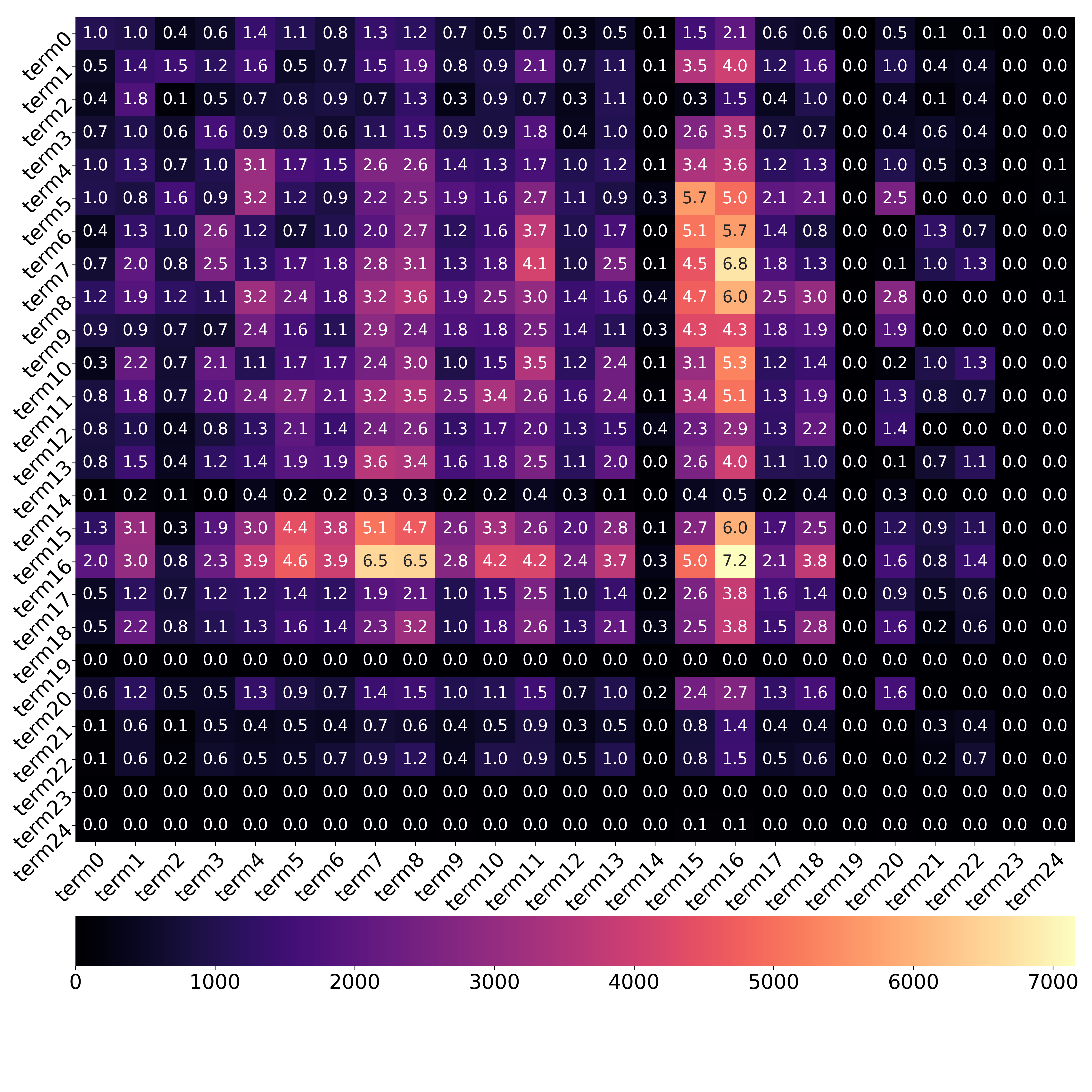}
    \caption{Kinship CP-N3 path scores ($\times 1000$).}
    \label{fig:kinshippathscores}
\end{figure}

One of the drawbacks of Mixture of Relation Paths (MP) is that \emph{only paths with relation path $\rvr$ contribute to the estimation of the weight parameter for $\rvr$ ($w_{\rvr}$}). In contrast, any path that contains $r_i$ in the $i^{th}$ edge of its relation path (a much larger set of paths) may contribute to the estimation of $w^i_{r_i}$ in \myeqnref{eqn:cm}. This is problematic because in most knowledge graphs, there is a stark difference in path counts for various relation paths. For example, \myfigref{fig:kinshipcounts} shows the counts for all length $2$ relation paths in Kinship and barring $4$ ({\tt term16}$(U,W)$ $\wedge$ {\tt term7}$(W,V)$, {\tt term16}$(U,W)$ $\wedge$ {\tt term8}$(W,V)$, {\tt term16}$(U,W)$ $\wedge$ {\tt term16}$(W,V)$, {\tt term7}$(U,W)$ $\wedge$ {\tt term16}$(W,V)$) all relation path counts lie below $1000$. In fact, a vast majority lie in single digits. This implies that for a rare relation path $\rvr$, estimated $w_{\rvr}$ may lack in statistical strength and thus may be unreliable.

\par

To address sparsity of paths, we utilize knowledge graph embeddings (KGE). The literature is rife with approaches that embed $\gV$ and $\gR$ into a low-dimensional hyperspace by learning distributed representations or \emph{embeddings}. Our assumption is that such techniques score \emph{more prevalent} paths \emph{higher} than paths that are less frequent. Let $\sigma(p)$ denote the score of path $p$ given such a pre-trained KGE. Score of $\langle u,r,v \rangle$ under the modified MP model is now given by:
\begin{equation*}
\sum_{\rvr =r_1, \ldots r_m} \text{LNN-pred}(\rve_{\rvr}) \sum_{p \in \gP_{\rvr}(u,v)} \sigma(p)
\end{equation*}
Essentially, the goal is to bias the learning process so that relation paths corresponding to larger $\sigma(p)$ are assigned larger weights. \myfigref{fig:kinshippathscores} shows the (sum of) path scores for all length $2$ relation paths in Kinship measured via CP-N3 embeddings \citep{lacroix2018canonical} where the cell values are much larger, e.g, while there are only $1200$ {\tt term16}$(U,W)$ $\wedge$ {\tt term16}$(W,V)$ paths its total path score is $7200$. Scoring paths using pre-trained KGE was only introduced recently in RNNLogic \citep{qu2020rnnlogic} which used RotatE \citep{sun2019rotate} specifically. We next show how to utilize a much larger class of KGEs for the same purpose. 
 
There are at least $2$ kinds of KGEs available that rely on a: 1) similarity measure of the triple $\langle h,r,t \rangle$, e.g., CP-N3 \citep{lacroix2018canonical}, and 2) distance measure used to contrast $t$'s embedding with some function of $h$ and $r$'s embeddings, e.g., TransE \citep{bordes:nips13}, RotatE \citep{sun2019rotate}. We describe $\sigma(p)$ for both cases:
\begin{align*}
\text{similarity-based: }\sigma(p)=&\sum_{\langle h,r,t \rangle \in p} \frac{1}{1 + \exp\{-\text{sim}(h, r, t)\}}\\
\text{distance-based: }\sigma(p)=&\sum_{\langle h,r,t \rangle \in p} \frac{\exp\{2(\delta - d(h, r, t))\} - 1}{\exp\{2(\delta - d(h, r, t))\} +1}
\end{align*}
where $\delta$ denotes the margin parameter used by the underlying distance-based KGE to convert distances into similarities. For both of these, we break the path into a series of edges, use the underlying KGE to compute similarity $\text{sim}()$ or distance $d()$ for each triple (as the case may be) and aggregate across all triples in the path. Based on extensive experimentation, we recommend {\tt sigmoid} and {\tt tanh} as the non-linear activation for similarity-based and distance-based KGE, respectively.

\subsection{Training Algorithm}

Among various possible training algorithms, based on extensive experimentation, we have found the following scheme to perform reliably. In each iteration, we sample uniformly at random a mini-batch of positive triples $B^+$ from $\langle h,r,t \rangle \in \gE$ and negative triples $B^-$ from $\langle h,r,t \rangle \notin \gE, \forall h,t \in \gV$, such that $|B^+| = |B^-|$ to minimize the following loss:
\begin{equation*}
\sum_{\langle h, r, t \rangle \in B^+} \sum_{\langle h^\prime, r, t^\prime \rangle \in B^-} \max\{0, \text{score}(h^\prime,t^\prime) - \text{score}(h,t) + \gamma\}
\end{equation*}
where $\gamma$ denotes the margin hyperparameter.

%% file: sections/experiments.tex
\section{Experimental Setup}
In this section, we describe our experimental setup. We introduce the datasets used and compare against state-of-the-art baselines. We primarily validate our LNN-based rule-learning approaches and their combination with KGE.  

\noindent \textbf{Datasets:} To evaluate our approach, we experiment on standard KBC benchmarks, viz. Unified Medical Language System (UMLS) \cite{kok2007statistical},  Kinship \cite{kok2007statistical}, WN18RR \cite{dettmers2018convolutional}, and FB15K-237 \cite{toutanova2015observed}. 
Table~\ref{tab:datasetstatistics} provides dataset statistics. 
We use standard train/validation/test splits for all datasets. 
Note that, RNNLogic defined its own splits for Kinship and UMLS, we report results on these too for a fair comparison (numbers in parenthesis in Table~\ref{tab:datasetstatistics}).

\eat{\noindent\textbf{Metrics:} In the recent past, there has been intense criticism on the metrics that are used to evaluate KBC techniques. Specifically, for the KBC task where answering $(h, r, ?)$ requires the method to assign probability to answers $t$, techniques, mostly in the rule-learning category, assigned same probability scores across multiple answers. The evaluation in most cases considered the minimum rank of the correct answer leading to overly optimistic and unfair evaluation~\cite{sun2019re}.  This has resulted in more accurate definition of KBC metrics proposed by \cite{sun2019re} which is used by RNNLogic\cite{qu2020rnnlogic}; the current state-of-the-art KBC approach. 
Specifically, \citet{sun2019re} proposed to compute the expectation over  all candidate entities with the same score as the correct answer. We compute the mean rank similar to ~\citet{qu2020rnnlogic}, i.e., $(m+ (n+1)/2)$, where $m$ is number of predictions scored higher and $n$ is the number of predictions that have the same score as the correct answer.}

\noindent \textbf{Baselines:} We compare our approaches against state-of-the-art KBC approaches categorized as follows:
\squishlist
\item \textit{Rule-learning approaches}: Neural-LP \cite{yang2017differentiable}, DRUM \cite{sadeghian2019drum}, RNNLogic (rules only) \cite{qu:iclr21}  and CTP \cite{minervini2020learning}. 
\item \textit{Rule-learning approaches that utilize KGE}: We compare against RNNLogic (w/ embd.) \cite{qu:iclr21} which uses RotatE \citep{sun:iclr19}, MINERVA \citep{das:iclr18} and MultiHopKG \citep{multihoplin2019}.
\item \textit{Embedding-based approaches}: We report results from ComplEx-N3, the best KGE we know of, and CP-N3, the KGE used by our methods, (both from \citet{lacroix2018canonical}), RotatE~\cite{sun:iclr19}, used in RNNLogic, and Complex \citep{trouillon2016complex}.

\squishend

\noindent\textbf{Metrics:} Given an unseen query $\langle h, r, ? \rangle$ we compute \emph{filtered} ranks \citep{bordes:nips13} for destination vertices after removing the destinations that form edges with $h$ and $r$ present in the train, test and validation sets. 
Based on \citet{sun:acl20}'s suggestions, our definitions of mean reciprocal rank (MRR) and Hits@K satisfy two properties: 1) They assign a larger value to destinations ranked lower, and 2) If destinations $t_1,\ldots,t_m$ share the same rank $n+1$ then each of them is assigned an average of ranks $n+1,\ldots, n+m$: 
\begin{align*}
\text{MRR}(t_i) = \frac{1}{m} \sum_{r=n+1}^{n+m} \frac{1}{r}, ~ \text{Hits@K}(t_i) = \frac{1}{m} \sum_{r=n+1}^{n+m} \delta(r \leq K)
\end{align*}
where $\delta()$ denotes the Dirac delta function. We include inverse triples and report averages across the test set. 

\noindent {\bf Implementation:} We evaluate rules learned with \emph{Chain of Mixtures} (LNN-CM), \emph{Mixture of Paths} (LNN-MP) and their combinations with KGE. We use Adagrad \citep{duchi:jmlr11} with step size $\in \{0.1, 1.0\}$, margin $\gamma \in \{0.1, 0.5, 1.0, 2.0\}$ and batch size $|B^+| = |B^-| = 8$. We use the validation set to perform hyperparameter tuning and learn rules of length up to $4$ for FB15K-237, $5$ for WN18RR, and $3$ for Kinship, UMLS. We combine our rule-learning approach with pre-trained CP-N3 embeddings of dimension $4K$ for FB15K-237, $3K$ for WN18RR, and $8K$ for Kinship, UMLS.

\begin{table}
\centering
{\scriptsize
\begin{tabular}{lccccc}
\toprule
Dataset & Train & Valid & Test & Relations & Entities \\
\midrule
\multirow{2}{*}{Kinship} & $8544 $ & $1068$ & $1074 $ & \multirow{2}{*}{$50$} & \multirow{2}{*}{$104$} \\
& $(3206)$ & $(2137)$ & $(5343)$ &&\\
\cmidrule{2-6}
\multirow{2}{*}{UMLS} & $5216$ & $652$ & $661$ & \multirow{2}{*}{$92$} & \multirow{2}{*}{$135$}\\
& $(1959)$ & $(1306)$ & $(3264)$ &&\\
\cmidrule{2-6}
WN18RR & $86835$ & $3034$ & $3134$ & $11$ & $40943$ \\
\cmidrule{2-6}
FB15K-237 & $272155$ & $17535$ & $20466$ & $237$ & $14541$ \\
\bottomrule
\end{tabular}
}
\caption{Dataset Statistics. Numbers in parenthesis reflects splits from RNNLogic~\cite{qu:iclr21}.}
\label{tab:datasetstatistics}
\end{table}

\begin{table*}
\centering
{\scriptsize
\begin{tabular}{llp{10pt}p{20pt}p{20pt}p{10pt}p{20pt}p{20pt}p{10pt}p{20pt}p{20pt}p{10pt}p{20pt}p{20pt}}
\toprule
&& \multicolumn{3}{c}{Kinship} & \multicolumn{3}{c}{UMLS} & \multicolumn{3}{c}{WN18RR} & \multicolumn{3}{c}{FB15K-237}\\
&& MRR & Hits@10 & Hits@3 & MRR & Hits@10 & Hits@3 & MRR & Hits@10 & Hits@3 & MRR & Hits@10 & Hits@3\\
\cmidrule(lr){3-5}
\cmidrule(lr){6-8}
\cmidrule(lr){9-11}
\cmidrule(lr){12-14}

\multirow{5}{*}{\rotatebox{90}{\textsc{Embd.}}}
& CP-N3 & $88.9$ & $\mathbf{98.7}$ & $\mathbf{94.8}$ & $93.3$ & $\mathbf{99.7}$ & $\mathbf{98.9}$  & $47.0^*$ & $54.0^*$ & $-$ &$36.0^*$&$54.0^*$&$-$\\
& Complex-N3 & $\mathbf{89.2}$ & $98.6$ & $94.7$ & $\mathbf{95.9}$ & $\mathbf{99.7}$ & $\mathbf{98.9}$  & $\mathbf{48.0}^*$ & $57.0^*$ & $-$ & $\mathbf{37.0}^*$ & $\mathbf{56.0}^*$ &  $-$\\
& RotatE & $75.5$ & $97.3$ & $86.5$ & $86.1$ & $99.5$ & $97.0$ & $47.6^*$ & $\mathbf{57.1}^*$ & $\mathbf{49.2}^*$ & $33.8^*$ & $53.3^*$ & $37.5^*$\\

& Complex & $83.7$ & $98.4$ & $91.6$ & $94.5$ & $\mathbf{99.7}$ & $98.0$ & $44.0$ & $49.6$ & $44.9$ & $31.8$ & $49.1$ & $34.7$ \\


\cmidrule{3-14}

\multirow{5}{*}{\rotatebox{90}{\textsc{Rules}}}& NeuralLP & $48.8$ & $89.1$ & $63.0$ & $55.3$ & $93.0$  & $75.4$  & $33.7$  & $50.2$  & $43.3$ & $25.8$&$48.5$  & $31.4$\\

& DRUM & $40.0$ & $86.1$ & $48.2$ & $61.8$ & $97.9$ & $91.2$  & $34.8$ & $52.1$ & $44.6$ &$25.8$&$\mathbf{49.1}$&$31.5$\\

& CTP & $70.3$ & $93.9$ & $79.7$ & $80.1$ & $97.0$ & $91.0$  & $-$ & $-$ & $-$ & $-$ & $-$ &  $-$\\
& RNNLogic & $64.5$ & $91.1$ & $72.9$ & $71.0$ & $91.1$ & $82.1$ & $45.5^*$ & $53.1^*$ & $47.5^*$ & $28.8^*$ & $44.5^*$ & $31.5^*$\\
& LNN-CM ({\bf Ours}) & $39.1$ & $68.7$ & $43.6$ & $78.7$ & $95.1$ & $88.9$ & $36.6$ & $49.2$ & $39.2$ & $28.0$ & $43.5$ & $30.4$\\
& LNN-MP ({\bf Ours}) & $\mathbf{81.9}$ & $\mathbf{98.4}$ & $\mathbf{89.3}$ & $\mathbf{90.0}$ & $\mathbf{99.4}$ & $\mathbf{98.3}$ & $\mathbf{47.3}$ & $\mathbf{55.5}$ & $\mathbf{49.7}$ & $\mathbf{30.7}$ & $47.0$ & $\mathbf{34.2}$\\
\cmidrule{3-14}
\multirow{4}{*}{\rotatebox{90}{\textsc{w/ Embd.}}}& RNNLogic & \multirow{2}{*}{$-$} & \multirow{2}{*}{$-$} & \multirow{2}{*}{$-$} & \multirow{2}{*}{$-$} & \multirow{2}{*}{$-$} & \multirow{2}{*}{$-$} & \multirow{2}{*}{$48.3^*$} & \multirow{2}{*}{$55.8^*$} & \multirow{2}{*}{$49.7^*$} & \multirow{2}{*}{$34.4^*$} & \multirow{2}{*}{$\mathbf{53.0}^*$} & \multirow{2}{*}{$38.0^*$}\\
& w/ RotatE & & & & & & & & & & & & \\
& LNN-MP ({\bf Ours}) & \multirow{2}{*}{$\mathbf{91.1}$} & \multirow{2}{*}{$\mathbf{99.2}$} & \multirow{2}{*}{$\mathbf{96.5}$} & \multirow{2}{*}{$\mathbf{94.5}$} & \multirow{2}{*}{$\mathbf{100}$} & \multirow{2}{*}{$\mathbf{99.2}$} & \multirow{2}{*}{$\mathbf{48.5}$} & \multirow{2}{*}{$\mathbf{56.1}$} & \multirow{2}{*}{$\mathbf{50.2}$} & \multirow{2}{*}{$\mathbf{35.1}$} & \multirow{2}{*}{$\mathbf{53.0}$} & \multirow{2}{*}{$\mathbf{39.1}$}\\
& w/ CP-N3 & & & & & & & & & & & & \\
\bottomrule
\end{tabular}
}
\caption{Results of LNN-CM and LNN-MP in comparison to other state-of-the-art approaches on standard splits of Kinship, UMLS, WN18RR, and FB15K-237. Bold font denotes best within each category of KBC approaches. $^*$ denotes results copied from original papers. CTP did not scale to larger datasets. ComplEx-N3 does not report Hits@3 for WN18RR and FB15K-237. RNNLogic does not report Kinship and UMLS results on standard splits w/ RotatE.}
\label{tab:results_standard_splits}
\end{table*}

\section{Results and Discussion}

\mytabref{tab:results_standard_splits} shows all our results comparing LNN-CM and LNN-MP on standard splits of all datasets. We compare against all the baselines including KGE-based approaches (\textsc{Embd.}), approaches that only learn rules (\textsc{Rules}), and approaches that learn rules with KGE (\textsc{w/ Embd.}). For the last category, we indicate the KGE used in the name of the approach.

\noindent{\bf LNN-CM vs. NeuralLP, DRUM}: Recall that LNN-CM is closely related to NeuralLP and DRUM. \mytabref{tab:results_standard_splits} shows that, in terms of MRR, LNN-CM outperforms DRUM and NeuralLP on UMLS and FB15K-237 while being comparable to DRUM on Kinship and WN18RR. LNN-CM achieves this without using an RNN-based rule generator and is thus conceptually much simpler. This is clear evidence that an RNN is not necessary for KBC. DRUM and NeuralLP's results in \mytabref{tab:results_standard_splits} are significantly lower than what was reported in \citet{yang:nips17,sadeghian:neurips19}. As explained in \citet{sun:acl20}, this is likely due to the carefully defined MRR and Hits@K used in our evaluation, not using which can lead to overly-optimistic and unfair results\footnote{Investigating more, Table 4 in \citeauthor{sadeghian:neurips19} reports DRUM's Hits@10 on WN18RR as $0.568$ with rule length up to 2. However, $56\%$ of WN18RR's validation set triples require paths longer than $2$ to connect the source with the destination (test set fraction is similar). This implies that one can only hope to achieve a maximum Hits@10 of $0.44$ when restricted to rules of length $2$ learned via \emph{any rule-based KBC} technique which is $< 0.568$ and a clear contradiction.}.


\noindent{\bf LNN-MP vs. LNN-CM}: \mytabref{tab:results_standard_splits} shows that LNN-MP (\textsc{Rules}) consistently outperforms LNN-CP. While previous comparisons against NeuralLP and/or DRUM \citep{qu:iclr21,multihoplin2019,das:iclr18} also hinted at this result, a surprising finding is that the margin of difference depends on the dataset. In particular, on  FB15K-237 LNN-CM's MRR comes within $1\%$ of  state-of-the-art RNNLogic (\textsc{Rules})’s. Across all standard splits, LNN-MP (\textsc{Rules}) outperforms all rule-learning methods. On the smaller Kinship and UMLS datasets, LNN-MP outperforms CTP, and on the larger WN18RR and FB15K-237 it outperforms RNNLogic. In \mytabref{tab:results_rnnlogic_splits}, we also report LNN-MP's results on RNNLogic's splits which include a much smaller training set for Kinship and UMLS (as shown in \mytabref{tab:datasetstatistics}). 
Given the substantial increases resulting from the switch to standard splits ($+14.5\%$ and $+8.5\%$ in MRR on Kinship and UMLS, respectively), it seems that the simpler LNN-MP (\textsc{Rules}) exploits the additional training data much more effectively. On the other hand, RNNLogic (\textsc{Rules}) hardly shows any improvement on Kinship and in fact, deteriorates on UMLS. A possible reason could be that the inexact training algorithm employed by RNNLogic (based on expectation-maximization and ELBO bound) fails to leverage the additional training data.
 

\noindent{\bf Learning Rules with Embeddings}: Since LNN-MP outperforms LNN-CM (rules only), we combine it with CP-N3, one of the best KGE available, to learn rules with embeddings. In comparison to LNN-MP (\textsc{Rules}), LNN-MP w/ CP-N3 shows consistent improvements ranging from $+1.2\%$ MRR (on WN18RR, \mytabref{tab:results_standard_splits}) to $+9.2\%$ MRR (on Kinship standard splits, \mytabref{tab:results_standard_splits}). In comparison to RNNLogic w/ RotatE, LNN-MP w/ CP-N3 shows small but consistent improvements on UMLS (RNNLogic's splits, \mytabref{tab:results_rnnlogic_splits}), WN18RR, FB15K-237 (\mytabref{tab:results_standard_splits}), and comparable results on Kinship (RNNLogic's splits, \mytabref{tab:results_rnnlogic_splits}). This is encouraging due to the simplicity of LNN-MP which adds to its appeal. \mytabref{tab:directtriples} shows that LNN-MP w/ CP-N3 also outperforms MINERVA and MultiHopKG, two more MP-based approaches that utilize embeddings. Note that, following MINERVA and MultiHopKG, for this comparison we leave out inverse triples from the test set. 



\begin{table}
\centering
{\scriptsize
\begin{tabular}{rp{10pt}p{15pt}p{15pt}p{10pt}p{15pt}p{15pt}}
\toprule
& \multicolumn{3}{c}{Kinship} & \multicolumn{3}{c}{UMLS}\\
& MRR & Hits@10 & Hits@3 & MRR & Hits@10 & Hits@3\\
\cmidrule(lr){2-4}
\cmidrule(lr){5-7}
CP-N3 & $60.2$ & $92.2$ & $70.0$ & $76.6$ & $95.0$ & $85.6$\\
ComplEx-N3 & $60.5^*$ & $92.1^*$ & $71.0^*$ & $79.1^*$ & $95.7^*$ & $87.3^*$\\
RotatE & $65.1^*$ & $93.2^*$ & $75.5^*$ & $74.4^*$ & $93.9^*$ & $82.2^*$\\

Complex & $54.4$ & $89.0$ & $63.8$ & $74.0$ & $92.5$ & $81.1$ \\


\cmidrule(lr){2-7}
RNNLogic & $63.9$ & $92.4$ & $73.1$ & $74.5$ & $92.4$ & $83.3$\\
LNN-MP ({\bf Ours}) & $\mathbf{67.4}$ & $\mathbf{95.0}$ & $\mathbf{77.0}$ & $\mathbf{81.5}$ & $\mathbf{96.6}$ & $\mathbf{91.8}$\\
\cmidrule(lr){2-7}
RNNLogic w/ RotatE & $\mathbf{72.2}^*$ & $94.9^*$ & $81.4^*$ & $84.2^*$ & $96.5^*$ & $89.1^*$\\
LNN-MP w/ CP-N3 ({\bf Ours}) & $\mathbf{72.2}$ & $\mathbf{96.9}$ & $\mathbf{81.6}$ & $\mathbf{87.6}$ & $\mathbf{98.0}$ & $\mathbf{94.7}$\\
\bottomrule
\end{tabular}
}
\caption{Results on RNNLogic's splits for Kinship and UMLS. $^*$ denotes results copied from \citet{qu:iclr21}.}
\label{tab:results_rnnlogic_splits}
\end{table}

\subsection{Qualitative Analysis and Discusssion} 

Learned rules can be extracted from LNN-MP by sorting the relation paths in descending order of $w_{\rvr}$. \mytabref{tab:learnedrules} presents some of the learned rules appearing in the top-10.

\begin{table}
\centering
{\scriptsize
\begin{tabular}{rp{10pt}p{15pt}p{15pt}p{10pt}p{15pt}p{15pt}}
\toprule
& \multicolumn{3}{c}{WN18RR} & \multicolumn{3}{c}{FB15K-237}\\
& MRR & Hits@10 & Hits@3 & MRR & Hits@10 & Hits@3\\
\cmidrule(lr){2-4}
\cmidrule(lr){5-7}
MINERVA$^*$ & $44.8$ & $51.3$ & $45.6$ & $29.3$ & $45.6$ & $32.9$\\
MultiHopKG$^*$ & $47.2$ & $54.2$ & $-$ & $40.7$ & $56.4$ & $-$ \\
RNNLogic$^*$ & $47.8$ & $55.3$ & $50.3$ & $37.7$ & $54.9$ & $41.2$\\
RNNLogic$^*$ w/ RotatE & $50.6$ & $\mathbf{59.2}$ & $52.3$ & $44.3$ & $\mathbf{64.0}$ & $48.9$\\
\cmidrule(lr){2-7}
LNN-MP ({\bf Ours}) & $51.6$ & $58.4$ & $53.5$ & $40.0$ & $57.4$ & $44.4$\\
LNN-MP w/ CP-N3 ({\bf Ours}) & $\mathbf{51.9}$ & $59.0$ & $\mathbf{53.9}$ & $\mathbf{44.8}$ & $63.6$ & $\mathbf{49.5}$\\
\bottomrule
\end{tabular}
}
\caption{Evaluating on direct triples (excluding inverses).}
\label{tab:directtriples}
\end{table}

\begin{table*}
\centering
{\scriptsize
\begin{tabular}{rl}
\toprule
\multirow{7}{*}{FB15K-237} & 1) $\text{person\_language}(P,L) \leftarrow \text{nationality}(P,N) \land~ \text{spoken\_in}(L,N)$\\
& 2) $\text{film\_language}(F,L) \leftarrow \text{film\_country}(F,C) \land~ \text{spoken\_in}(L,C)$\\
& 3) $\text{tv\_program\_language}(P,L) \leftarrow \text{country\_of\_tv\_program}(P,N) \land~ \text{official\_language}(N,L)$\\
& 4) $\text{burial\_place}(P,L) \leftarrow \text{nationality}(P,N) \land~ \text{located\_in}(L,N)$\\
& 5) $\text{country\_of\_tv\_program}(P,N) \leftarrow \text{tv\_program\_actor}(P,A) \land~ \text{born\_in}(A, L) \land~ \text{located\_in}(L,N)$\\
& 6) $\text{film\_release\_region}(F,R) \leftarrow \text{film\_crew}(F,P) \land~ \text{marriage\_location}(P,L) \land~ \text{located\_in}(L,R)$\\
& 7) $\text{marriage\_location}(P,L) \leftarrow \text{celebrity\_friends}(P,F) \land~ \text{marriage\_location}(F,L^\prime) \land~ \text{location\_adjoins}(L^\prime,L)$\\
\midrule
\multirow{3}{*}{WN18RR} & 8) $\text{domain\_topic}(X,Y) \leftarrow \text{hypernym}(X,U) \land~ \text{hypernym}(U,V) \land~ \text{hypernym}(W,V) \land~ \text{hypernym}(Z,W) \land~ \text{domain\_topic}(Z,Y)$\\
& 9) $\text{derivation}(X,Y) \leftarrow \text{hypernym}(X,U) \land~ \text{hypernym}(V,U) \land~\text{derivation}(W,V) \land~\text{hypernym}(W,Z) \land~\text{hypernym}(Y,Z)$\\
& 10) $\text{hypernym}(X,Y) \leftarrow \text{member\_meronym}(U,X) \land~ \text{hypernym}(U,V) \land~ \text{hypernym}(W,V) \land~ \text{member\_meronym}(W,Z) \land~ \text{hypernym}(Z,Y)$\\
\bottomrule
\end{tabular}
}
\caption{Examples of LNN-MP's learned rules.}
\label{tab:learnedrules}
\end{table*}

\noindent{\bf Rules for FB15K-237}: Rule 1 in \mytabref{tab:learnedrules} describes a learned rule that infers the language a person speaks by exploiting knowledge of the language spoken in her/his country of nationality. In terms of relation paths, this looks like: 
\centerline{$P (\text{person}) \overset{\text{nationality}}{\longrightarrow} N (\text{nation}) \overset{\text{spoken\_in}}{\longrightarrow} L (\text{language})$}\\
Similarly, Rule 2 uses the $\text{film\_country}$ relation instead of $\text{nationality}$ to infer the language used in a film. Besides $\text{spoken\_in}$, FB15K-237 contains other relations that can be utilized to infer language such as the $\text{official\_language}$ spoken in a country. Rule 3 uses this relation to infer the language spoken in a TV program by first exploiting knowledge of its country of origin. Rules 5, 6 and 7 are longer rules containing $3$ relations each in their body. Rule 5 infers a TV program's country by first exploiting knowledge of one of its actor's birth place and then determining which country the birth place belongs to. In terms of relation path: $P (\text{program}) \overset{\text{tv\_program\_actor}}{\longrightarrow} A (\text{actor}) \overset{\text{born\_in}}{\longrightarrow} L (\text{birth place})$ $\overset{\text{located\_in}}{\longrightarrow} N (\text{nation})$. Rule 6 uses a film crew member's marriage location instead to infer the region where the film was released. Rule 7 infers the marriage location of a celebrity by exploiting knowledge of where their friend got married.

\noindent{\bf Recursive Rules for WN18RR}: Unlike FB15K-237, WN18RR is a hierarchical KG and thus sparser which calls for longer rules to be learned. A majority of WN18RR's edges are concentrated in a few relations, e.g. $\text{hypernym}$, and most learned rules rely on such relations. $\text{Hypernym}(X,Y)$ is true if $Y$, e.g. dog, is a kind of $X$, e.g. animal. \mytabref{tab:learnedrules} presents $3$ learned rules for WN18RR. $\text{Domain\_topic}(X,Y)$ is true if $X$ is the category, e.g., computer science, of scientific concept $Y$, e.g., CPU. Rule 8 infers the domain topic of $Y$ by first climbing down the hypernym is-a hierarchy 2 levels and then climbing up: 
$X \overset{\text{is-a}}{\rightarrow} U \overset{\text{is-a}}{\rightarrow} V \overset{\text{is-a}}{\leftarrow} W \overset{\text{is-a}}{\leftarrow} Z \overset{\text{domain}}{\longrightarrow} Y$. Rules 9 and 10 are recursive, where the relation in the head of the rule also appears in the body. Here, $\text{derivation}$ indicates related form of concept, e.g., yearly is the derivation of year, and $\text{member\_meronym}$ denotes a concept which is a member of another concept such as player and team.

%% file: sections/related_work_2.tex
\section{Related Work}

Knowledge Base Completion (KB) approaches have gained a lot of interest recently, due to their ability to handle the incompleteness of knowledge bases. 
Embedding-based techniques maps entities and relations to low-dimensional vector space to infer new facts. These techniques use neighborhood structure of an entity or relation to learn their corresponding embedding. Starting off with translational embeddings~\cite{bordes2013translating}, embedding-based techniques have evolved to use complex vector spaces such as ComplEx~\cite{trouillon2016complex}, RotatE~\cite{sun2019rotate}, and QuatE~\cite{zhang2019quaternion}. 
While the performance of embedding-based techniques has improved over time, rule-learning techniques have gained attention due to its inherent ability for learning interpretable rules ~\cite{yang2017differentiable,sadeghian2019drum,rocktaschel2017end, qu2020rnnlogic}. 

The core ideas in rule learning can be categorized into two groups based on their mechanism to select relations for rules. While \textit{Chain  of  Mixtures (CM)} represents each relation in the body as a mixture, e.g. NeuralLP~\cite{yang:nips17}, DRUM~\cite{sadeghian:neurips19}, \textit{Mixture  of Paths (MP)} learns relation paths; e.g., MINERVA~\cite{das:iclr18}, RNNLogic~\cite{qu:iclr21}. Recent trends in both these types of rule learning approaches has shown significant increase in complexity for performance gains over their simpler precursors. 
Among the first to learn rules for KBC, NeuralLP \citep{yang:nips17} uses a long short-term memory (LSTM) \citep{hochreiter:neuralcomp97} as its rule generator. DRUM \citep{sadeghian:neurips19} improves upon NeuralLP by learning multiple such rules obtained by running a bi-directional LSTM for more steps. MINERVA \citep{das:iclr18} steps away from representing each body relation as a mixture, proposing instead to learn the relation sequences appearing in paths connecting source to destination vertices using neural reinforcement learning. 

RNNLogic \citep{qu:iclr21} is the latest to adopt a path-based approach for KBC that consists of two modules, a rule-generator for suggesting high quality paths and a reasoning predictor that uses said paths to predict missing information. RNNLogic employs expectation-maximization for training where the E-step identifies useful paths per data instance (edge in the KG) by sampling from an intractable posterior while the M-step uses the per-instance useful paths to update the overall set of paths. Both DRUM and RNNLogic represent a significant increase in complexity of their respective approaches compared to NeuralLP and MINERVA. 

Unlike these approaches, we propose to utilize \emph{Logical Neural Networks} (LNN) \citep{riegel:arxiv20}; a simple yet powerful neuro-symbolic approach which extends Boolean logic to the real-valued domain. On top of LNN, we propose two simple approaches for rule learning based on CP and MP. Furthermore, our approach allows for combining  rule-based KBC with any KGE, which results in state-of-the-art performance across multiple datasets. 

%% file: sections/conclusion.tex
\section{Conclusion}

In this paper, we proposed an approach for rule-based KBC based on Logic Neural Networks (LNN); a neuro-symbolic differentiable framework for real-valued logic. In particular, we present two approaches for rule learning, one that represent rules using chains of predicate mixtures (LNN-CM), and another that uses mixtures of paths (LNN-MP). We show that both approaches can be implemented using LNN and neuro-symbolic AI, and result in better or comparable performance to state-of-the-art. Furthermore,  our framework facilitates combining rule learning with knowledge graph embedding techniques to harness the best of both worlds. 

Our experimental results across four benchmarks show that such a combination provides better results and establishes new state-of-the-art performance across multiple datasets. 
We also showed how easy it is to extract and interpret learned rules. 
With both LNN-CM and LNN-MP implemented within the same LNN framework, one avenue of future work would be to explore combinations of the two approaches given their good performance on a number of KBC benchmarks.

%% file: kbc.bbl
\begin{thebibliography}{31}
\providecommand{\natexlab}[1]{#1}

\bibitem[{Bollacker et~al.(2008)Bollacker, Evans, Paritosh, Sturge, and
  Taylor}]{bollacker2008freebase}
Bollacker, K.; Evans, C.; Paritosh, P.; Sturge, T.; and Taylor, J. 2008.
\newblock Freebase: a collaboratively created graph database for structuring
  human knowledge.
\newblock In \emph{Proceedings of the 2008 ACM SIGMOD international conference
  on Management of data}, 1247--1250.

\bibitem[{Bordes et~al.(2013{\natexlab{a}})Bordes, Usunier, Garcia-Duran,
  Weston, and Yakhnenko}]{bordes:nips13}
Bordes, A.; Usunier, N.; Garcia-Duran, A.; Weston, J.; and Yakhnenko, O.
  2013{\natexlab{a}}.
\newblock Translating embeddings for modeling multi-relational data.
\newblock In \emph{NeurIPS}.

\bibitem[{Bordes et~al.(2013{\natexlab{b}})Bordes, Usunier, Garcia-Duran,
  Weston, and Yakhnenko}]{bordes2013translating}
Bordes, A.; Usunier, N.; Garcia-Duran, A.; Weston, J.; and Yakhnenko, O.
  2013{\natexlab{b}}.
\newblock Translating embeddings for modeling multi-relational data.
\newblock \emph{Advances in neural information processing systems}, 26.

\bibitem[{Das et~al.(2018)Das, Dhuliawala, Zaheer, Vilnis, Durugkar,
  Krishnamurthy, Smola, and McCallum}]{das:iclr18}
Das, R.; Dhuliawala, S.; Zaheer, M.; Vilnis, L.; Durugkar, I.; Krishnamurthy,
  A.; Smola, A.; and McCallum, A. 2018.
\newblock Go for a Walk and Arrive at the Answer: Reasoning Over Paths in
  Knowledge Bases using Reinforcement Learning.
\newblock In \emph{ICLR}.

\bibitem[{Dettmers et~al.(2018)Dettmers, Minervini, Stenetorp, and
  Riedel}]{dettmers2018convolutional}
Dettmers, T.; Minervini, P.; Stenetorp, P.; and Riedel, S. 2018.
\newblock Convolutional 2d knowledge graph embeddings.
\newblock In \emph{Thirty-second AAAI conference on artificial intelligence}.

\bibitem[{Dong et~al.(2019)Dong, Mao, Lin, Wang, Li, and Zhou}]{dong:iclr19}
Dong, H.; Mao, J.; Lin, T.; Wang, C.; Li, L.; and Zhou, D. 2019.
\newblock Neural Logic Machines.
\newblock In \emph{ICLR}.

\bibitem[{Duchi, Hazan, and Singer(2011)}]{duchi:jmlr11}
Duchi, J.; Hazan, E.; and Singer, Y. 2011.
\newblock Adaptive Subgradient Methods for Online Learning and Stochastic
  Optimization.
\newblock \emph{JMLR}.

\bibitem[{Frerix, Nießner, and Cremers(2020)}]{frerix:cvprw20}
Frerix, T.; Nießner, M.; and Cremers, D. 2020.
\newblock Homogeneous Linear Inequality Constraints for Neural Network
  Activations.
\newblock In \emph{CVPR Workshops}.

\bibitem[{Hochreiter and Schmidhuber(1997)}]{hochreiter:neuralcomp97}
Hochreiter, S.; and Schmidhuber, J. 1997.
\newblock Long short-term memory.
\newblock \emph{Neural computation}.

\bibitem[{Kok and Domingos(2007)}]{kok2007statistical}
Kok, S.; and Domingos, P. 2007.
\newblock Statistical predicate invention.
\newblock In \emph{Proceedings of the 24th international conference on Machine
  learning}, 433--440.

\bibitem[{Lacroix, Usunier, and Obozinski(2018)}]{lacroix2018canonical}
Lacroix, T.; Usunier, N.; and Obozinski, G. 2018.
\newblock Canonical tensor decomposition for knowledge base completion.
\newblock In \emph{International Conference on Machine Learning}, 2863--2872.
  PMLR.

\bibitem[{Lao, Mitchell, and Cohen(2011)}]{lao:emnlp11}
Lao, N.; Mitchell, T.; and Cohen, W.~W. 2011.
\newblock Random Walk Inference and Learning in A Large Scale Knowledge Base.
\newblock In \emph{EMNLP}.

\bibitem[{Lin, Socher, and Xiong(2018)}]{multihoplin2019}
Lin, X.~V.; Socher, R.; and Xiong, C. 2018.
\newblock Multi-hop knowledge graph reasoning with reward shaping.
\newblock In \emph{EMNLP}.

\bibitem[{Minervini et~al.(2020)Minervini, Riedel, Stenetorp, Grefenstette, and
  Rockt{\"a}schel}]{minervini2020learning}
Minervini, P.; Riedel, S.; Stenetorp, P.; Grefenstette, E.; and
  Rockt{\"a}schel, T. 2020.
\newblock Learning reasoning strategies in end-to-end differentiable proving.
\newblock In \emph{International Conference on Machine Learning}, 6938--6949.
  PMLR.

\bibitem[{Pascanu, Mikolov, and Bengio(2013)}]{pascanu:icml13}
Pascanu, R.; Mikolov, T.; and Bengio, Y. 2013.
\newblock On the Difficulty of Training Recurrent Neural Networks.
\newblock In \emph{ICML}.

\bibitem[{Qu et~al.(2020)Qu, Chen, Xhonneux, Bengio, and Tang}]{qu2020rnnlogic}
Qu, M.; Chen, J.; Xhonneux, L.-P.; Bengio, Y.; and Tang, J. 2020.
\newblock Rnnlogic: Learning logic rules for reasoning on knowledge graphs.
\newblock \emph{arXiv preprint arXiv:2010.04029}.

\bibitem[{Qu et~al.(2021)Qu, Chen, Xhonneux, Bengio, and Tang}]{qu:iclr21}
Qu, M.; Chen, J.; Xhonneux, L.-P.; Bengio, Y.; and Tang, J. 2021.
\newblock {\{}RNNL{\}}ogic: Learning Logic Rules for Reasoning on Knowledge
  Graphs.
\newblock In \emph{ICLR}.

\bibitem[{Rebele et~al.(2016)Rebele, Suchanek, Hoffart, Biega, Kuzey, and
  Weikum}]{rebele2016yago}
Rebele, T.; Suchanek, F.; Hoffart, J.; Biega, J.; Kuzey, E.; and Weikum, G.
  2016.
\newblock YAGO: A multilingual knowledge base from wikipedia, wordnet, and
  geonames.
\newblock In \emph{International semantic web conference}, 177--185. Springer.

\bibitem[{Riegel et~al.(2020)Riegel, Gray, Luus, Khan, Makondo, Akhalwaya,
  Qian, Fagin, Barahona, Sharma, Ikbal, Karanam, Neelam, Likhyani, and
  Srivastava}]{riegel:arxiv20}
Riegel, R.; Gray, A.; Luus, F.; Khan, N.; Makondo, N.; Akhalwaya, I.~Y.; Qian,
  H.; Fagin, R.; Barahona, F.; Sharma, U.; Ikbal, S.; Karanam, H.; Neelam, S.;
  Likhyani, A.; and Srivastava, S. 2020.
\newblock Logical Neural Networks.
\newblock \emph{CoRR}.

\bibitem[{Rockt{\"a}schel and Riedel(2017)}]{rocktaschel2017end}
Rockt{\"a}schel, T.; and Riedel, S. 2017.
\newblock End-to-end differentiable proving.
\newblock \emph{arXiv preprint arXiv:1705.11040}.

\bibitem[{Sadeghian et~al.(2019{\natexlab{a}})Sadeghian, Armandpour, Ding, and
  Wang}]{sadeghian:neurips19}
Sadeghian, A.; Armandpour, M.; Ding, P.; and Wang, D.~Z. 2019{\natexlab{a}}.
\newblock DRUM: End-To-End Differentiable Rule Mining On Knowledge Graphs.
\newblock In \emph{NeurIPS}.

\bibitem[{Sadeghian et~al.(2019{\natexlab{b}})Sadeghian, Armandpour, Ding, and
  Wang}]{sadeghian2019drum}
Sadeghian, A.; Armandpour, M.; Ding, P.; and Wang, D.~Z. 2019{\natexlab{b}}.
\newblock Drum: End-to-end differentiable rule mining on knowledge graphs.
\newblock \emph{arXiv preprint arXiv:1911.00055}.

\bibitem[{Sun et~al.(2019{\natexlab{a}})Sun, Deng, Nie, and
  Tang}]{sun2019rotate}
Sun, Z.; Deng, Z.-H.; Nie, J.-Y.; and Tang, J. 2019{\natexlab{a}}.
\newblock Rotate: Knowledge graph embedding by relational rotation in complex
  space.
\newblock \emph{arXiv preprint arXiv:1902.10197}.

\bibitem[{Sun et~al.(2019{\natexlab{b}})Sun, Deng, Nie, and Tang}]{sun:iclr19}
Sun, Z.; Deng, Z.-H.; Nie, J.-Y.; and Tang, J. 2019{\natexlab{b}}.
\newblock RotatE: Knowledge Graph Embedding by Relational Rotation in Complex
  Space.
\newblock In \emph{ICLR}.

\bibitem[{Sun et~al.(2020)Sun, Vashishth, Sanyal, Talukdar, and
  Yang}]{sun:acl20}
Sun, Z.; Vashishth, S.; Sanyal, S.; Talukdar, P.; and Yang, Y. 2020.
\newblock A Re-evaluation of Knowledge Graph Completion Methods.
\newblock In \emph{ACL}.

\bibitem[{Toutanova and Chen(2015)}]{toutanova2015observed}
Toutanova, K.; and Chen, D. 2015.
\newblock Observed versus latent features for knowledge base and text
  inference.
\newblock In \emph{Proceedings of the 3rd workshop on continuous vector space
  models and their compositionality}, 57--66.

\bibitem[{Trinh et~al.(2018)Trinh, Dai, Luong, and Le}]{trinh:iclrw18}
Trinh, T.~H.; Dai, A.~M.; Luong, M.-T.; and Le, Q.~V. 2018.
\newblock Learning Longer-term Dependencies in RNNs with Auxiliary Losses.
\newblock In \emph{ICLR Workshops}.

\bibitem[{Trouillon et~al.(2016)Trouillon, Welbl, Riedel, Gaussier, and
  Bouchard}]{trouillon2016complex}
Trouillon, T.; Welbl, J.; Riedel, S.; Gaussier, {\'E}.; and Bouchard, G. 2016.
\newblock Complex embeddings for simple link prediction.
\newblock In \emph{International conference on machine learning}, 2071--2080.
  PMLR.

\bibitem[{Yang, Yang, and Cohen(2017{\natexlab{a}})}]{yang:nips17}
Yang, F.; Yang, Z.; and Cohen, W.~W. 2017{\natexlab{a}}.
\newblock Differentiable Learning of Logical Rules for Knowledge Base
  Reasoning.
\newblock In \emph{NeurIPS}.

\bibitem[{Yang, Yang, and Cohen(2017{\natexlab{b}})}]{yang2017differentiable}
Yang, F.; Yang, Z.; and Cohen, W.~W. 2017{\natexlab{b}}.
\newblock Differentiable learning of logical rules for knowledge base
  reasoning.
\newblock \emph{arXiv preprint arXiv:1702.08367}.

\bibitem[{Zhang et~al.(2019)Zhang, Tay, Yao, and Liu}]{zhang2019quaternion}
Zhang, S.; Tay, Y.; Yao, L.; and Liu, Q. 2019.
\newblock Quaternion knowledge graph embeddings.
\newblock \emph{arXiv preprint arXiv:1904.10281}.

\end{thebibliography}
